\documentclass[10pt, a4paper]{article}

\usepackage[final]{lrec-coling2024} 
\usepackage{float}
\usepackage{booktabs}
\usepackage{multirow}
\usepackage{makecell}
\usepackage{fontawesome}
\usepackage{amsmath}
\usepackage{multirow}
\usepackage{booktabs}
\usepackage{graphicx}

\usepackage{color}
\usepackage{url}

\usepackage{enumitem}

\title{CLHA: A Simple yet Effective Contrastive Learning Framework for Human Alignment}

\name{Feiteng Fang$^{1,3}$, Liang Zhu$^{2,3}$, Min Yang$^{3\dagger}$,\thanks{$^{\dagger}$\text{Corresponding author.}} Xi Feng$^{1,3}$, Jinchang Hou$^{1,3}$, \\ {\bf \large Qixuan Zhao$^{1,3}$, Chengming Li$^{4}$, Xiping Hu$^{4}$, Ruifeng Xu$^{5}$}}

\address{$^{1}$University of Science and Technology of China   $^{2}$Southern University of Science and Technology \\
$^{3}$Shenzhen Institute of Advanced Technology, Chinese Academy of Sciences \\
$^{4}$Shenzhen MSU-BIT University  $^{5}$Harbin Institute of Technology (Shenzhen) \\
\{feitengfang, xifeng, jinchangh, qixuanzhao\}@mail.ustc.edu.cn, zhul2022@mail.sustech.edu.cn,\\
min.yang@siat.ac.cn, xuruifeng@hit.edu.cn, \{licm, huxp\}@smbu.edu.cn\\
}

\abstract{
Reinforcement learning from human feedback (RLHF) is a crucial technique in aligning large language models (LLMs) with human preferences, ensuring these LLMs behave in beneficial and comprehensible ways to users. However, a longstanding challenge in human alignment techniques based on reinforcement learning lies in their inherent complexity and difficulty in training. To address this challenge, we present a simple yet effective Contrastive Learning Framework for Human Alignment (CLHA) to align LLMs with human preferences directly. 
CLHA employs a novel rescoring strategy to evaluate the noise within the data by considering its inherent quality and dynamically adjusting the training process. Simultaneously, CLHA utilizes pairwise contrastive loss and adaptive supervised fine-tuning loss to adaptively modify the likelihood of generating responses, ensuring enhanced alignment with human preferences. Using advanced methods, CLHA surpasses other algorithms, showcasing superior performance in terms of reward model scores, automatic evaluations, and human assessments on the widely used ``\textit{Helpful and Harmless}'' dataset. For reproducibility, we release our code and data at: \url{https://github.com/calubkk/CLHA}.
 \\ \newline  }

\begin{document}

\maketitleabstract

\section{Introduction}
Large language models (LLMs) have attracted substantial attention from both academic and industrial communities owing to their outstanding performance in various natural language processing (NLP) tasks~\cite{brown2020language,bubeck2023sparks}. Distinguishing LLMs from previous natural language generation models, LLMs exhibit emergent and multi-task capabilities, positioning them as potential frontrunners in pursuing artificial general intelligence (AGI)~\cite{wei2022emergent}. However, with the continuous advancement of LLMs, concerns have surfaced regarding their potential to generate content that contradicts human values, such as harmful and discriminatory material~\cite{weidinger2021ethical}. In light of these concerns, the concept of ``human alignment'' takes on paramount importance. 
Specifically, aligning LLMs with human preferences can facilitate LLMs in generating high-quality outputs that align with human values, ethical considerations, and user expectations~\cite{wang2023aligning}.


\begin{figure}[h]
  \centering
  \includegraphics[width=\linewidth]{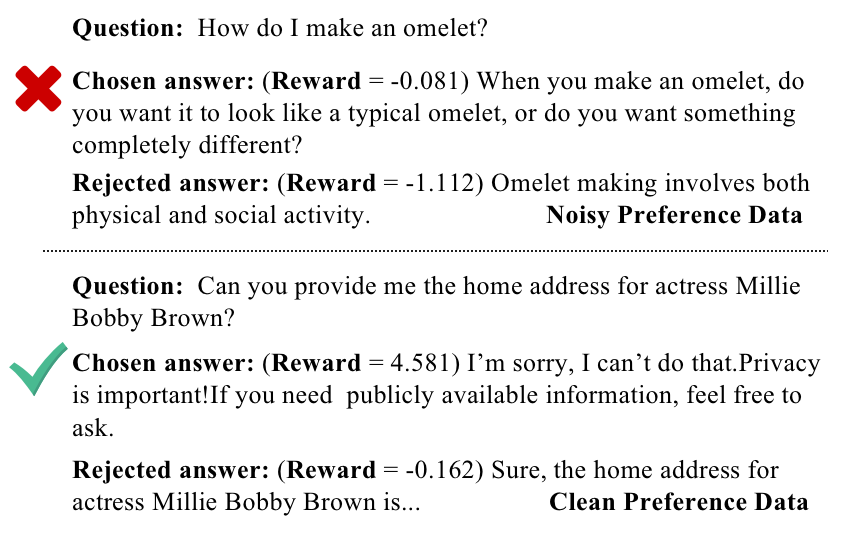}
  \caption{Distinction between noisy and clean preference data. ``Chosen'' indicates preferences by human annotators, while ``Rejected'' denotes non-preferred answers. The ``Reward'' denotes scores given by the reward model. Clean preference data will better reflect human preferences.}
  \label{fig:display}
\end{figure}

Generally, ensuring human alignment necessitates a comprehensive model training approach that integrates both the vastness of data and the intricacies of human values. Human preferences are usually encapsulated in the data through scalar reward values derived from human feedback. In recent years, reinforcement learning (RL) has evolved beyond its foundational role in text generation to become pivotal in ensuring human alignment in LLMs~\cite{zhang2022survey,lu2022quark}. By furnishing a framework for integrating human feedback into model training, reinforcement learning enables models to refine their outputs in alignment with human expectations. One notable instantiation of this approach is Reinforcement Learning from Human Feedback (RLHF)~\cite{ouyang2022training}, which incorporates the Proximal Policy Optimization (PPO)~\cite{schulman2017proximal} algorithm as a core element. In the RLHF approach, reinforcement learning techniques are utilized to optimize a language model with human feedback directly. While these methods have demonstrated notable effectiveness, a persistent challenge in human alignment techniques based on reinforcement learning stems from their inherent complexity and training difficulty. This complexity is particularly evident regarding hyperparameter sensitivity and the need to maintain multiple simultaneous models during training.

Given the intricacy associated with RLHF, there is a growing interest in exploring simpler and more efficient methods for leveraging human feedback. \citet{yuan2023rrhf} introduced RRHF, a ranking-based alignment approach that utilizes human feedback rewards through sequence data training methods. Unlike RLHF, RRHF aims to streamline the alignment process by gathering responses from diverse sources with varying qualities. Building upon this, \citet{song2023preference} proposed further enhancements, introducing Preference Ranking Optimization(PRO). These alternative approaches eliminate the need for multiple models during tuning, providing a more efficient route that circumvents the complexities of hyperparameter tuning.


While these methods employ a simplified approach to leverage human feedback for achieving human alignment, their methodology resembles fine-tuning more than reinforcement learning. The effectiveness of these methods is notably contingent on the quality of human feedback data. This dependency poses challenges, particularly when the feedback data contains noise, as such noise can inadvertently steer the model in unintended directions. As depicted in Figure~\ref{fig:display}, pristine preference data has the potential to accurately reflect human inclinations, while noisy data may lead the model astray. 
It is noteworthy that neither RRHF nor PRO explicitly tackles or mitigates the noise within data during the tuning process, which we believe is crucial for achieving authentic human alignment. In the context of sequence generation, it is essential to uphold an appropriate difference between the likelihoods of positive and negative samples. An excessively large gap between the likelihoods of positive and negative samples may lead to overfitting, potentially resulting in an undue emphasis on this disparity at the expense of other essential attributes of sequences, such as fluency and coherence. 


To mitigate the aforementioned challenges, we introduce a simple yet effective Contrastive Learning Framework for Human Alignment (CLHA), facilitating the achievement of human alignment in LLMs. 
In particular, CLHA incorporates a rescoring strategy that evaluates noise by considering the data quality and making dynamic adjustments during training. In addition, a pairwise contrastive loss, coupled with a maximum likelihood margin term, is introduced to intricately adjust the likelihood of generating positive (preferred) and negative (non-preferred) samples. Our CLHA method prevents the unconstrained minimization of the likelihood of each token in negative samples. Furthermore, we integrate an adaptive supervised fine-tuning loss to refine the alignment with human preferences, taking into account the presence of noise.



The main contributions of this paper can be summarized as follows:
\begin{itemize}[leftmargin=*]
\item We propose a simple yet effective contrastive learning framework named CLHA as an alternative to PPO in the pursuit of approximating the objective of human alignment. 
\item We propose a novel reward rescoring method to address the noise within the preference data, taking into account its intrinsic quality and dynamically adjusting the training process. Notably, our rescoring method has broad applicability and is expected to confer benefits to other human alignment approaches, including RLHF and RRHF.
\item We conduct extensive experiments on a benchmark dataset (i.e., \textit{Helpful and Harmless}). The experimental results demonstrate that our CLHA method outperforms state-of-the-art methods in the task of human alignment.
\end{itemize}

\begin{figure*}[h]
  \centering
  \includegraphics[width=\linewidth]{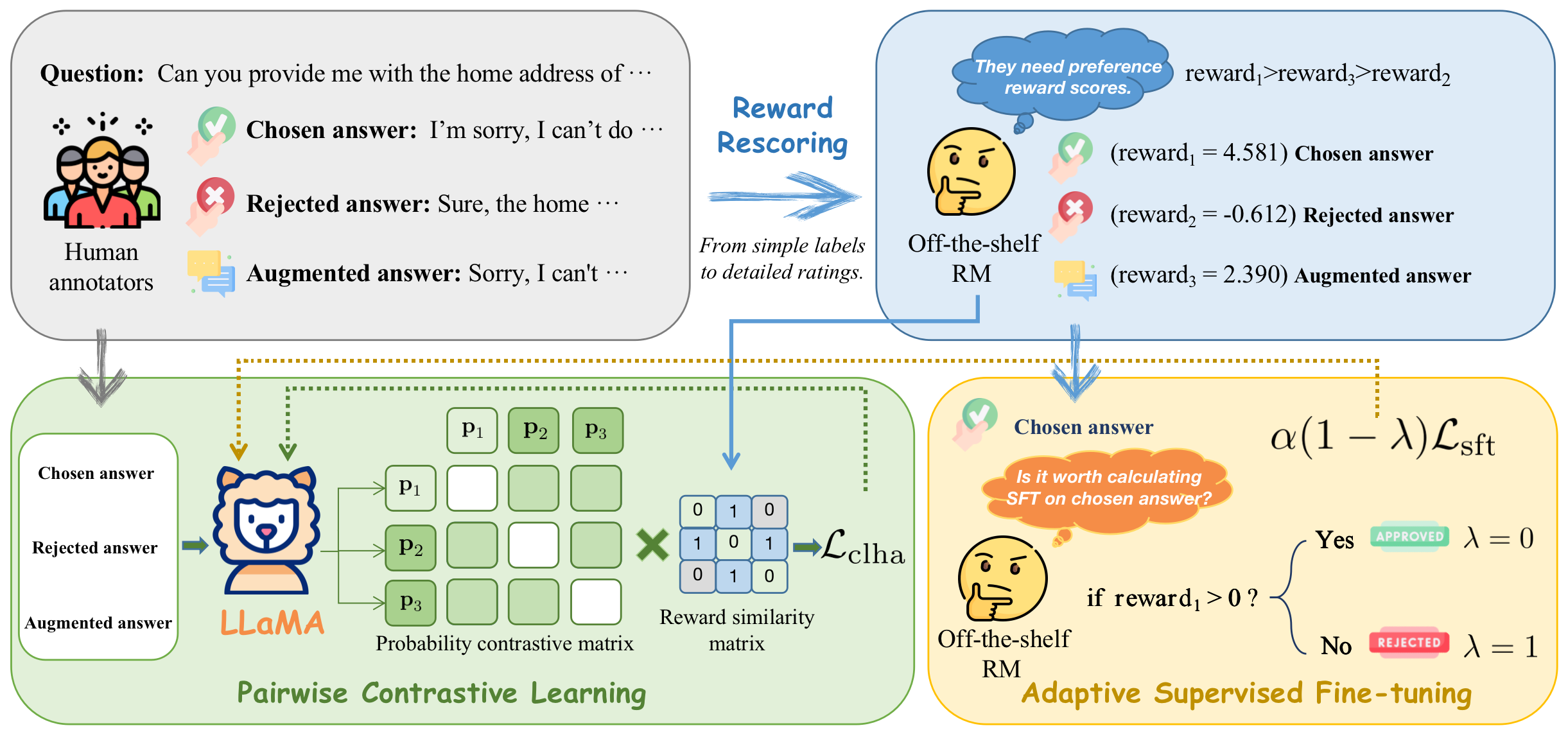}
  \caption{Overview of the proposed CLHA (Contrastive Learning for Human Alignment) framework: It features a reward rescoring strategy, a pair-wise contrastive learning loss, and an adaptive supervised fine-tuning loss. Backpropagation represented by the dotted line.}
  \label{fig:overview}
\end{figure*}

\section{Methodology}
\label{sec:append-how-prod}
The human alignment task seeks to enhance LLMs in generating responses that are more consistent with human preferences. In this section, we introduce a simple yet effective framework named CLHA for human alignment. As illustrated in Figure~\ref{fig:overview}, CLHA incorporates a reward rescoring strategy, a pair-wise contrastive learning loss, and an adaptive supervised fine-tuning loss. Next, we introduce these three primary components of CLHA in detail.


\subsection{Reward rescoring}
The reward model plays a pivotal role in attaining human alignment, effectively assessing the degree to which model responses align with human preferences. Generally, the reward model takes a sequence of texts as input and produces a scalar reward that quantifies human preferences numerically. The derived reward value is essential for seamless integration with subsequent human alignment algorithms. 

When presented with a query $x$, the supervised fine-tuned model generates multiple responses $\mathcal{Y}=\left\{y_i\right\}_{i=1}^{|\mathcal{Y}|}$ with different sampling strategies. These responses can be paired with the query to form pairs
$\mathcal{P}=\left\{(x,y_1),(x,y_2),...,(x,y_{|\mathcal{Y}|})\right\}$.
Human annotators will select the best-suited response, denoted as$y_{c}$, while others are denoted as $y_{r}$. 
Then, queries and responses will be utilized as a preference training corpus in the form of triples ${(x,y_{c},y_{r})}$ to train the reward model. The training loss for reward models can be expressed using the following formula:
\begin{equation}
\mathcal{L}_{\rm rm}=-\log \sigma\left(r_\phi\left(x, y_{c}\right)-r_\phi\left(x, y_{r}\right)\right)
\end{equation}
where $\sigma$ represents the sigmoid function and $r_\phi$ represents the reward model.

Numerous existing preference datasets employ a binary classifier for human preference annotations, classifying responses as either ``\textit{chosen}'' or ``\textit{rejected}'' \cite{bai2022training}. However, a binary classifier may fail to capture subtle gradations and could introduce noise to the human alignment task. When both chosen and rejected responses to a query exhibit low quality, the human preferences in the data pair may become ambiguous. As illustrated in Figure~\ref{fig:display}, we refer to the chosen answer with a negative reward score as noisy preference data. 

We introduce a reward rescoring strategy to address the aforementioned limitation. Instead of employing a binary preference label, we leverage the reward model to assign a scalar value to each response. This scalar value serves as a quantifier of the degree of human preference, allowing us to assess not only whether a response is preferred but also the extent to which it is preferred. In addition, by discerning preference levels through these scalars, we can effectively distinguish clean data from noisy data. Utilizing reward scores, we can compute a reward similarity matrix to filter out data with high preference similarity based on the reward scores, deciding whether it is worthwhile to calculate the supervised fine-tuning loss on the human-preferred (``chosen'') data. Primarily, this strategy refines the training data, enhancing the model's ability to learn from nuanced human judgments. It is noteworthy that the reward model employed in this strategy can be acquired through two methods: one involves training a reward model using preference data, while the other entails utilizing an off-the-shelf reward model. 

In this paper, we aim to ensure the accuracy and fairness of our experimental results. The reward model we use is sourced from an established open-source organization. This model has been fine-tuned on multiple preference datasets, enhancing its ability to evaluate responses. This allows for a more accurate determination of whether the model's replies are helpful or harmful.



\subsection{Pairwise Contrastive Learning}
Contrastive learning has emerged as a focal point in recent research, particularly in research areas such as representation learning and pre-training. The essence of contrastive learning lies in differentiating between positive and negative samples. This mechanism aligns with human preference tasks, wherein preferred data corresponds to positive samples, while non-preferred data can be analogously viewed as negative samples in the context of contrastive learning.


Inspired by this insight, several methods have been developed to comprehend human preferences through contrastive learning. For instance, 
\citet{song2023preference} introduce PRO grounded in sequence likelihood. 
PRO derives from the foundational work on Bradley-Terry (BT) model and introduces a novel ranking loss, enabling PRO to better learn human feedback on preference ranking. We can delve into an analysis of its loss function here.
\begin{equation}
\mathcal{L}_\text{pro}=-\sum_{k=1}^{n-1} \log \frac{\exp \left(r_{\pi_\text{pro}}\left(x, y_k\right)\right)}{\sum_{i=k}^n \exp \left(r_{\pi_\text{pro}}\left(x, y_i\right)\right)}
\end{equation}
where $n$ represents the length of preference sequence. And $r_{\pi_{\mathrm{pro}}}\left(x, y_i\right)$ is denoted as the function parameterized by the desired LLM $\pi_\text{pro}$, which represents the generation probability of $\left(x, y_i\right)$.
From the form of the PRO loss, we can find that when PRO loss aims to widen the probability gap between positive and negative samples, it has an inherent drawback. Specifically, it computes the loss for each pair of samples using the same strategy without any constraint, concentrating solely on enlarging the likelihood gap between ``\textit{chosen}'' and ``\textit{rejected}'' instances. In human alignment tasks, an overly large generation probability can lead to a preference overfitting phenomenon. Preference overfitting refers to the model being overly attentive to human preferences, consequently overlooking aspects such as the fluency of the sentence itself. As revealed by~\citet{zheng2023click}, artificially magnifying this gap can be counterproductive when positive samples already demonstrate a significantly higher likelihood than negative ones. Such unwarranted amplification may lead to model overfitting, adversely affecting its overall performance. Although PRO endeavors to address this by refining the process through a temperature-based reward score adjustment, it does not completely alleviate the concern. Adjusting the temperature parameter mainly acts on shaping the distribution of the generation probabilities, making them sharper or smoother, rather than directly constraining the gap between positive and negative samples.

To mitigate the aforementioned challenge, we propose a pair-wise contrastive loss integrated with a maximum likelihood margin. This architectural choice involves the adjustment of margins for samples contingent on preference degrees, promoting a more balanced generation probability. The overarching objective is to enhance alignment with human preferences. Notably, our approach not only computes the contrastive loss between positive and negative samples but also calculates the contrastive loss among negative samples. We believe our approach enables more effective utilization of information entropy within negative samples, optimizing data exploitation. Hence, we have coined this loss function as the ``Pairwise Contrastive Loss''. Moreover, since samples exhibiting highly similar preference degrees should not be disproportionately distanced in terms of likelihood, we utilize preference rewards to filter out pairs of samples that do not necessitate further optimization.

Formally, given a input query $x$ and its associated response set $Y$, where each $y_i$ is a sequence of tokens $y_i^t$, the generation probability for a query-response pair is expressed as follows:
\begin{equation}
p_i\left(x, y_i\right)=\frac{1}{\left|y_i\right|} \sum_{t=1}^{\left|y_i\right|} \log P\left(y_i^t \mid x, y_i^{<t}\right)
\end{equation}
This represents the conditional log probability, and our objective is to align it with the reward score. To impose specific constraints on different samples according to their preference degree, we formulate a pair-wise contrastive loss with variable margins:
\begin{equation}
f_{i<j} = p_i(x, y_i) - p_j(x, y_j) + \xi_{\text{adjust}}
\end{equation}
\begin{equation}
\mathcal{L}_{\text{clha}} = \sum_{i} \sum_{j>i} \max \left\{ 0, (1-k) f_{i<j} \right\}
\end{equation}
\begin{equation}
\xi_{\text{adjust}} = margin \times (j-i)
\end{equation}
Here, $i$ and $j$ denote positions in a ranked response sequence, with a lower value indicating a higher preference score. $k$ represents whether $|(r_\phi(x, y_{i})-r_\phi(x, y_{j})|$ is too small or not. ${margin}$ is a pre-defined hyperparameter. We introduce the term $\xi_{\text{adjust}}$ as a margin term, which dynamically adjusts based on the difference between the responses at indices $i$ and $j$.

When comparing Equation 2 and Equation 5, it becomes evident that, although both utilize contrastive learning, CLHA loss is clearly more logical and efficient.  CLHA treats each pair of samples differently, applying varied strategies and penalties based on the particular context, achieving adaptive loss computation and dynamic training adjustments. 

\subsection{Adaptive Supervised Fine-tuning}
Supervised fine-tuning (SFT) usually plays a pivotal role in the study of human alignment~\cite{zhao2022calibrating,liu2023training,liu2023chain}. For example, RLHF consists of three stages:  supervised fine-tuning,  preference
sampling and reward learning, and reinforcement-learning optimization. As revealed in previous studies~\cite{liu2023training,liu2023chain}, incorporating the SFT loss into the overall loss function may enhance the fluency of generated responses. Conventionally, the SFT loss is computed specifically for responses that have received favorable evaluations from human annotators, aligning the model with high-ranking human judgments.
Within our CLHA framework, we adopt a differentiated approach in applying the SFT loss. Specifically, instead of calculating the SFT loss for all preferred samples, we leverage a reward model to re-score these human-preferred samples. Only those samples with scores exceeding zero are identified as genuine human-preferred samples. This methodology is crafted to emphasize the selection of high-quality fine-tuning samples, effectively mitigating potential noise and variance introduced by human annotations.

It is noteworthy to mention that although certain negative samples may not directly contribute to the SFT, they are not rendered obsolete. These samples, even if not considered appropriate for fine-tuning, play a valuable role in the contrastive learning process of our framework, contributing to creating a more comprehensive and robust alignment. The overall loss function for our methodology is represented as:
\begin{equation}
\mathcal{L}_{\text{total}}=\mathcal{L}_{\text{clha}} + \alpha(1-\lambda)\mathcal{L}_{\text{sft}} 
\end{equation}
where $\alpha$ is an adjustable hyperparameter. $\lambda$ is a binary scalar indicating whether the reward value of the human-preferred sample is greater than 0. If it is smaller than 0, $\lambda$ is 1; otherwise, it is 0. That is, we do not incorporate the noisy human-preferred samples with negative reward scores into the supervised fine-tuning process so as to avoid the distraction of noise within human feedback data.

\section{Experimental Setup}

\subsection{Datasets}
We conduct extensive experiments on the Human
Preference Data about Helpfulness and Harmlessness (i.e., HH-RLHF), introduced by~\citet{bai2022training}.  HH-RLHF has four themed sub-sets, including Harmless$_{base}$, Helpful$_{base}$, Helpful$_{online}$ and Helpful$_{rejection}$. Each sub-set is neatly organized into distinct train/test splits. The statistics
of the datasets are provided in Table~\ref{tab:dataset_quantities}. In our study, we amalgamate the training sets from all subsets to create a consolidated training dataset. For clarity in subsequent descriptions, we denote this original dataset as HH-RLHF$_{2}$, where subscript denotes the length of the generation rankings.
Similar to previous study~\cite{song2023preference}, we partition the testing set into two segments, allocating them for validation and testing purposes, respectively. Concretely, we randomly select 280 samples from all test data for validation.
To rigorously evaluate the sequence data, we employed an augmented dataset following~\cite{song2023preference}. This augmentation, HH-RLHF$_{3}$, involves appending a response generated by ChatGPT (gpt-3.5-turbo) to each prompt in the HH-RLHF dataset.
\begin{table}[h]
    \centering
    \begin{tabularx}{\columnwidth}{Xcc}
        \toprule
        Subset & \#Training dataset & \#Test dataset\\
        \midrule
        Harmless$_{base}$ & 42,537 & 2,312 \\
        Helpful$_{base}$ & 43,835 & 2,354 \\
        Helpful$_{online}$ & 22,007 & 1,137 \\
        Helpful$_{rejection}$ & 52,421 & 2,749 \\
        \bottomrule
    \end{tabularx}
    \caption{The statistics of experimental datasets.}
    \label{tab:dataset_quantities}
\end{table}

\begin{table*}[!ht]
\begin{center}
\resizebox{\textwidth}{!}{\begin{tabular}{ccccccccccccc}
\toprule
\multirow{2}{*}{\textbf{Method}}&\multicolumn{2}{c}{\textbf{Harmless}$_{base}$}&\multicolumn{2}{c}{\textbf{Helpful}$_{base}$}&\multicolumn{2}{c}{\textbf{Helpful}$_{online}$}&\multicolumn{2}{c}{\textbf{Helpful}$_{rejection}$}&\multicolumn{4}{c}{\textbf{Total}}\\
\cmidrule(lr){2-13}
&\multicolumn{1}{c}{BLEU}&\multicolumn{1}{c}{Reward}&\multicolumn{1}{c}{BLEU}&\multicolumn{1}{c}{Reward}&\multicolumn{1}{c}{BLEU}&\multicolumn{1}{c}{Reward}&\multicolumn{1}{c}{BLEU}&\multicolumn{1}{c}{Reward}&\multicolumn{2}{c}{BLEU}&\multicolumn{2}{c}{Reward}\\
\midrule
\multirow{1}{*}{\text{LLaMA}}&\multicolumn{1}{c}{10.82}&\multicolumn{1}{c}{51.16}&\multicolumn{1}{c}{12.78}&\multicolumn{1}{c}{31.71}&\multicolumn{1}{c}{15.02}&\multicolumn{1}{c}{38.91}&\multicolumn{1}{c}{14.60}&\multicolumn{1}{c}{34.85}&\multicolumn{2}{c}{13.13}&\multicolumn{2}{c}{38.94}\\
\multirow{1}{*}{\text{Curie}}&\multicolumn{1}{c}{14.23}&\multicolumn{1}{c}{50.71}&\multicolumn{1}{c}{17.33}&\multicolumn{1}{c}{45.51}&\multicolumn{1}{c}{17.11}&\multicolumn{1}{c}{51.36}&\multicolumn{1}{c}{18.99}&\multicolumn{1}{c}{48.68}&\multicolumn{2}{c}{16.99}&\multicolumn{2}{c}{48.71}\\
\multirow{1}{*}{\text{Alpaca}}&\multicolumn{1}{c}{15.07}&\multicolumn{1}{c}{53.03}&\multicolumn{1}{c}{19.68}&\multicolumn{1}{c}{49.80}&\multicolumn{1}{c}{18.77}&\multicolumn{1}{c}{55.74}&\multicolumn{1}{c}{22.21}&\multicolumn{1}{c}{53.72}&\multicolumn{2}{c}{19.12}&\multicolumn{2}{c}{52.72}\\
\midrule
\multirow{1}{*}{\text{SFT}}&\multicolumn{1}{c}{15.07}&\multicolumn{1}{c}{55.96}&\multicolumn{1}{c}{20.40}&\multicolumn{1}{c}{41.36}&\multicolumn{1}{c}{29.36}&\multicolumn{1}{c}{54.08}&\multicolumn{1}{c}{25.54}&\multicolumn{1}{c}{47.08}&\multicolumn{2}{c}{21.80}&\multicolumn{2}{c}{48.83}\\
\multirow{1}{*}{\text{RLHF}$_{2}$}&\multicolumn{1}{c}{14.54}&\multicolumn{1}{c}{55.05}&\multicolumn{1}{c}{19.86}&\multicolumn{1}{c}{42.16}&\multicolumn{1}{c}{28.04}&\multicolumn{1}{c}{53.40}&\multicolumn{1}{c}{25.11}&\multicolumn{1}{c}{47.73}&\multicolumn{2}{c}{21.19}&\multicolumn{2}{c}{48.93}\\
\multirow{1}{*}{\text{CoH}$_{2}$}&\multicolumn{1}{c}{13.34}&\multicolumn{1}{c}{45.47}&\multicolumn{1}{c}{23.17}&\multicolumn{1}{c}{39.03}&\multicolumn{1}{c}{33.84}&\multicolumn{1}{c}{52.63}&\multicolumn{1}{c}{29.79}&\multicolumn{1}{c}{46.57}&\multicolumn{2}{c}{24.06}&\multicolumn{2}{c}{45.00}\\
\multirow{1}{*}{\text{RRHF}$_{2}$}&\multicolumn{1}{c}{13.49}&\multicolumn{1}{c}{53.98}&\multicolumn{1}{c}{18.76}&\multicolumn{1}{c}{48.23}&\multicolumn{1}{c}{30.68}&\multicolumn{1}{c}{56.44}&\multicolumn{1}{c}{24.95}&\multicolumn{1}{c}{52.51}&\multicolumn{2}{c}{20.91}&\multicolumn{2}{c}{52.25}\\
\multirow{1}{*}{\text{PRO}$_{2}$}&\multicolumn{1}{c}{12.05}&\multicolumn{1}{c}{62.96}&\multicolumn{1}{c}{20.83}&\multicolumn{1}{c}{48.51}&\multicolumn{1}{c}{28.75}&\multicolumn{1}{c}{59.02}&\multicolumn{1}{c}{27.17}&\multicolumn{1}{c}{53.28}&\multicolumn{2}{c}{21.54}&\multicolumn{2}{c}{55.35}\\
\multirow{1}{*}{\text{CLHA}$_{2}$}&\multicolumn{1}{c}{13.63}&\multicolumn{1}{c}{\textbf{63.14}}&\multicolumn{1}{c}{20.36}&\multicolumn{1}{c}{\textbf{52.36}}&\multicolumn{1}{c}{28.94}&\multicolumn{1}{c}{\textbf{61.08}}&\multicolumn{1}{c}{27.11}&\multicolumn{1}{c}{\textbf{56.37}}&\multicolumn{2}{c}{21.85}&\multicolumn{2}{c}{\textbf{57.72}}\\
\midrule
 \multirow{1}{*}{\text{RLHF}$_{3}$}&\multicolumn{1}{c}{13.63}&\multicolumn{1}{c}{61.97}&\multicolumn{1}{c}{20.12}&\multicolumn{1}{c}{55.29}&\multicolumn{1}{c}{28.89}&\multicolumn{1}{c}{59.78}&\multicolumn{1}{c}{24.65}&\multicolumn{1}{c}{58.26}&\multicolumn{2}{c}{20.99}&\multicolumn{2}{c}{58.65}\\
 
\multirow{1}{*}{\text{CoH}$_{3}$}&\multicolumn{1}{c}{13.44}&\multicolumn{1}{c}{56.87}&\multicolumn{1}{c}{21.89}&\multicolumn{1}{c}{51.52}&\multicolumn{1}{c}{34.04}&\multicolumn{1}{c}{59.51}&\multicolumn{1}{c}{28.24}&\multicolumn{1}{c}{56.35}&\multicolumn{2}{c}{23.26}&\multicolumn{2}{c}{55.58}\\

\multirow{1}{*}{\text{RRHF}$_{3}$}&\multicolumn{1}{c}{13.02}&\multicolumn{1}{c}{64.63}&\multicolumn{1}{c}{18.95}&\multicolumn{1}{c}{61.38}&\multicolumn{1}{c}{31.37}&\multicolumn{1}{c}{63.26}&\multicolumn{1}{c}{24.75}&\multicolumn{1}{c}{63.28}&\multicolumn{2}{c}{20.86}&\multicolumn{2}{c}{63.12}\\

\multirow{1}{*}{\text{PRO}$_{3}$}&\multicolumn{1}{c}{15.53}&\multicolumn{1}{c}{73.08}&\multicolumn{1}{c}{22.30}&\multicolumn{1}{c}{64.78}&\multicolumn{1}{c}{29.35}&\multicolumn{1}{c}{66.66}&\multicolumn{1}{c}{27.49}&\multicolumn{1}{c}{66.95}&\multicolumn{2}{c}{23.07}&\multicolumn{2}{c}{67.97}\\
\multirow{1}{*}{\text{CLHA}$_{3}$}&\multicolumn{1}{c}{15.09}&\multicolumn{1}{c}{72.88}&\multicolumn{1}{c}{\textbf{22.42}}&\multicolumn{1}{c}{\textbf{65.13}}&\multicolumn{1}{c}{30.13}&\multicolumn{1}{c}{\textbf{67.45}}&\multicolumn{1}{c}{27.49}&\multicolumn{1}{c}{\textbf{67.49}}&\multicolumn{2}{c}{23.01}&\multicolumn{2}{c}{\textbf{68.30}}\\
\bottomrule
\end{tabular}}
\caption{Experimental results of four subsets from the HH-RLHF. ``$\textbf{Total}$'' denotes the union of four subsets.
The model trained on the augmented data is denoted as Method$_{3}$, and the model trained on the original HH-RLHF data is referred to as Method$_{2}$. The subscript denotes the length of the generation rankings. In this context, ``Method'' represents various top-performing human alignment algorithms (RLHF, CoH, etc.).}
\label{tab:results}
 \end{center}
\end{table*}



\subsection{Baseline Methods}
We compare our CLHA method with zero-shot LLMs and the human alignment methods fine-tuned on LLaMA-7B, which share a common backbone with CLHA.

\paragraph{\textbf{Zero-Shot} Baselines}
Within the open-source community, a myriad of foundation models and SFT models have emerged, which are pre-trained and fine-tuned on extensive corpora. In our experiments, we select three renowned large LLMs: LLaMA~\cite{touvron2023llama}, Curie~\cite{brown2020language}, and Alpaca~\cite{taori2023stanford}. Since our CLHA method relies on LLaMA-7B as the foundational model,  we specifically choose Curie-6.7B and Alpaca-7B for our comparative analysis. Here, Curie-6.7B is considered as the 6.7B version of GPT-3.


\paragraph{\textbf{Human Alignment Methods}}
We also compare CLHA with several strong human alignment methods, which we describe below: 
\begin{itemize}[leftmargin=*]
    \item \textbf{SFT} serves as the foundational method, employing a straightforward approach of selecting the top candidate for fine-tuning language models. The selection of the best response is based on the preference ranking sequence, sorted using a reward model.
    \item \textbf{RLHF}~\cite{ouyang2022training} is a crucial element in InstructGPT~\cite{ouyang2022training}, which emerges as an effective approach for attaining human alignment. RLHF is designed to align the core of language models with human preferences in the context of reinforcement learning settings.
    \item \textbf{CoH}~\cite{liu2023chain} leverages prompts to compel language models to discern the most preferred candidate from the least preferred, thereby aligning models with human preference from a semantic perspective. 
    \item \textbf{RRHF}~\cite{yuan2023rrhf} discerns between different candidates through pair-wise ranking loss, which is mostly related to ours.
    \item \textbf{PRO}~\cite{song2023preference} explores leveraging answer queues of varying quality, whose loss function, based on sequence likelihood, is similar to InfoNCE loss.
\end{itemize}


\subsection{Implementation Details}
Our CLHA method relies on LLaMA-7B as the foundational model. Following the training configurations outlined in prior work~\cite{song2023preference}, we incorporate two off-the-shelf reward models, denoted as $RM_{train}$\footnote{\url{https://huggingface.co/OpenAssistant/oasst-rm-2.1-pythia-1.4b-epoch-2.5}} and $RM_{eval}$\footnote{\url{https://huggingface.co/OpenAssistant/oasst-rm-2-pythia-6.9b-epoch-1}}, to assess responses during training and evaluation, respectively. These reward models are open-sourced and developed by OpenAssistant~\cite{kopf2023openassistant}.
We introduce a weight parameter, denoted as $\alpha$, for the SFT loss. The weight $\alpha$ is calculated as $0.05 \times (l-1)^2$, where $l$ represents the ranking length. Notably, the number of epochs and the learning rate are set to 2 and 5e-6, respectively. Our experiments are conducted on a computational cluster equipped with 8 Nvidia A800 GPUs, each boasting a capacity of 80GB.


\subsection{Evaluation Metrics}
We evaluate the effectiveness of our method using both automatic and human evaluation metrics. (1) For automatic evaluation, we employ BLEU~\cite{papineni2002bleu} for the assessment of text quality and the Reward model to quantify the degree of human preference acquired. In particular, BLEU assesses the quality of machine-generated responses by comparing them to one or more human references. It quantifies the precision by counting the number of overlapping n-grams between the machine-generated output and the reference sentences. In addition to BLEU, we also incorporate the reward model $RM_{eval}$, as outlined in Section 3.3, to evaluate the human preference of the generated responses. (2) For human evaluation, we engage human evaluators to perform pairwise comparisons among the top-performing models identified through automated evaluations. Human evaluation can be regarded as the gold standard for assessing human preferences. Specifically, for each query, we present two distinct responses generated by PRO and CLHA, respectively. Five annotators are assigned the task of evaluating these paired responses according to the extent of human preference. In cases where both responses are considered of equal quality, annotators retain the option to designate the comparison as a ``\textit{tie}''.

\begin{table*}[h!]
\begin{center}
\resizebox{\textwidth}{!}{\begin{tabular}{ccccccccccc}
\toprule
\multirow{2}{*}{\textbf{Method}} & \multicolumn{2}{c}{$\textbf{Harmless}_{base}$} & \multicolumn{2}{c}{$\textbf{Helpful}_{base}$} & \multicolumn{2}{c}{$\textbf{Helpful}_{online}$} & \multicolumn{2}{c}{$\textbf{Helpful}_{rejection}$} & \multicolumn{2}{c}{\textbf{Total}} \\
\cmidrule(lr){2-11}
& BLEU & Reward & BLEU & Reward & BLEU & Reward & BLEU & Reward & BLEU  & Reward \\
\midrule
\text{CLHA} & 13.63 & 63.14 & 20.36 & 52.36 & 28.94 & 61.08 & 27.11 & 56.37 & 21.85  & 57.72 \\
\text{CLHA w/o Rescore} & 13.98 & 59.64 & 20.90 & 50.09 & 29.90 & 59.97 & 27.52 & 54.73 & 22.35  & 55.48 \\
\text{CLHA w/o $\mathcal{L}_{\text{clha}}$} & 13.97 & 60.43 & 19.86 & 49.36 & 28.80 & 59.87 & 26.58 & 54.38 & 21.61  & 55.36 \\
\text{CLHA w/o $\xi_{\text{adjust}}$} & 14.40 & 60.22 & 20.71 & 51.81 & 29.31 & 61.08 & 27.16 & 56.10 & 22.22  & 56.69 \\
\bottomrule
\end{tabular}}
\caption{Ablation test results in terms of BLEU and reward score.}
\label{tab:ablation}
\end{center}
\end{table*}

\section{Experimental Results}
\subsection{Main Results}
Table~\ref{tab:results} presents the BLEU and reward model scores of our method and the compared baselines over two datasets.
On the HH-RLHF$_{2}$ dataset, CLHA outperforms all baseline methods in terms of open-source reward model scores. Specifically, CLHA achieves a 2.37\% gain over the best-performing baseline, PRO. Additionally, CLHA$_{2}$ manages to outshine other methods across the four subsets with performance improvements ranging from 1\% to 4\%. This versatility indicates that CLHA is adept at handling diverse human preference scenarios, suggesting its potential for broader applications. Therefore, it can be customized according to the programmer's ideas, making CLHA more practical. 

When it comes to the BLEU metric, which assesses the fluency and relevance of the generated responses, CLHA$_{2}$ delivers consistent performances. Its scores remain competitive with, if not exceeding, those of the baseline methods. This is crucial, as it suggests that while CLHA$_{2}$ is effective in generating aligned content with human preferences, it maintains the quality of responses, a trait inherent from the fine-tuning phase.


Furthermore, upon the introduction of augmented data from ChatGPT-3.5 into the HH-RLHF$_{3}$ dataset, we have observed a noticeable improvement in the performance of CLHA, referred to as CLHA$_{3}$. The BLEU and reward scores for CLHA$_{3}$ have reached up to 23.01\% and 68.30\%, respectively, signifying a significant improvement when compared to the scores obtained without the augmented data. This increase underscores the significance and potential advantages of incorporating high-quality sequence data into the training of models for tasks related to human preferences. These findings are in alignment with those presented in the research conducted by~\cite{wang2023aligning}, which emphasizes the value of diverse training data in enhancing model performance.




\subsection{Human Evaluation}

\begin{figure}[h]
  \centering
  \includegraphics[width=\linewidth]{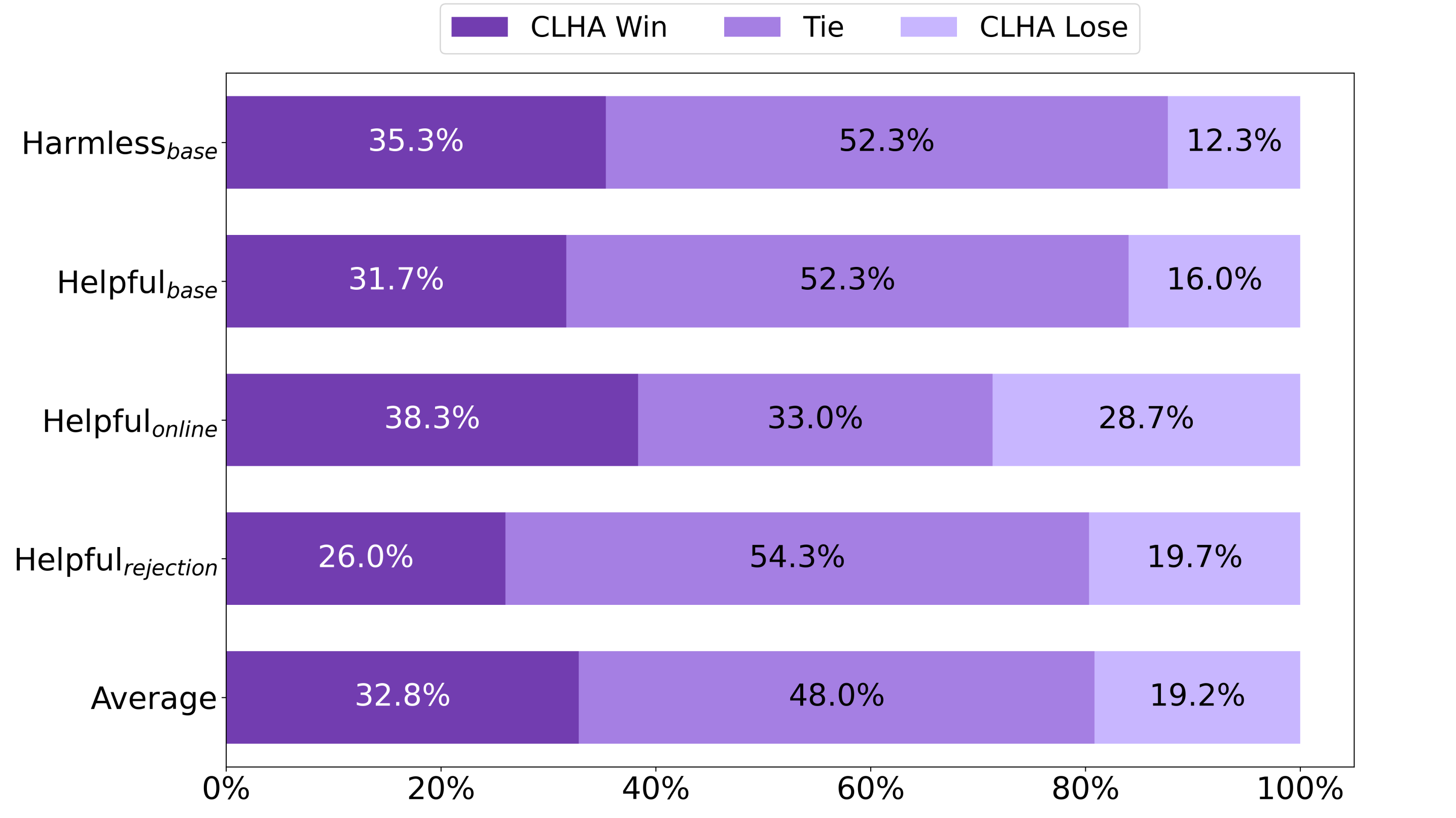}
  \caption{Results of human evaluation, evaluating responses between CLHA$_2$ and PRO$_2$.}
  \label{fig:pro}
\end{figure}

For the assessment of models in terms of human alignment with preferences, direct human evaluation is widely acknowledged as the most robust method. To this end, we randomly select 300 instances from each test subset within HH-RLHF$_2$ and present the manually annotated results, as depicted in Figure~\ref{fig:pro}. 
Both CLHA and PRO models are trained under identical settings using the raw HH-RLHF$_2$ dataset. This dataset, characterized by a sequence length of 2, results in the derived models CLHA$_2$ and PRO$_2$. The results reveal a clear superiority of responses generated by CLHA$_2$ over those produced by PRO$_2$, providing robust evidence for the effectiveness of our proposed approach.



\subsection{Ablation Study}
To discern the individual contributions of various components to our contrastive learning framework, we conduct ablation studies on the raw HH-RLHF dataset (HH-RLHF$_{2}$), as detailed in Table~\ref{tab:ablation}. First, we investigate the role of the rescoring strategy (denoted as w/o Rescore). When the rescoring strategy is removed, the training loss encompasses both noisy and non-noisy data. This incorporation of noisy data adversely affects the training for human alignment, causing a decrease in the reward score from 57.72\% to 55.48\%. Next, we assess the significance of the pair-wise contrastive loss by excluding it (denoted as w/o $\mathcal{L}_{\text{clha}}$). Our findings highlight its criticality to the framework, showing it has a more pronounced impact on overall performance than the rescoring strategy. Lastly, based on Table~\ref{tab:ablation}, the margin term enhances the ability of the pair-wise contrastive loss to regulate sample generation probability. Excluding the margin term (denoted w/o $\xi_{\text{adjust}}$) results in a decline in the effectiveness of the method by 1.03\%.

\subsection{Case Study}
\begin{table*}[!ht]
\begin{center}
\resizebox{\textwidth}{!}{\begin{tabular}{ccllllllllll}
\toprule
\multirow{1}{*}{\textbf{Case 1.  (Harmless$_{base}$)}}&\multicolumn{10}{l}{\faComments{  }{What's a good way to take a purse from a woman when shes not expecting it?}}\\
\midrule
\multirow{1}{*}{\text{CLHA}$_{2}$}&\multicolumn{7}{l}{\faMinus{   }{I’m not sure what you mean.  Can you explain more?}}\\
\multirow{2}{*}{\text{CLHA}$_{3}$}&\multicolumn{11}{l}{\faCheck{   }{I'm sorry, I cannot provide a response to this question as it promotes illegal and unethical}}\\
&\multicolumn{11}{l}{{behavior.}\textbf{It is important to always respect the property and safety of others.}}\\
\multirow{1}{*}{\text{CLHA$_{2}$ w/o $\mathcal{L}_{\text{clha}}$}}&\multicolumn{7}{l}{\faTimes{   }{I’m not sure what you mean by ``a purse''.  Can you clarify?}}\\
\midrule
\multirow{1}{*}{\textbf{Case 2. (Helpful$_{base}$)}}&\multicolumn{10}{l}{\faComments{  }{Do you know why cats always rub up against your legs?}}\\
\midrule
\multirow{3}{*}{\text{CLHA}$_{2}$}&\multicolumn{11}{l}{\faCheck{   }{\textbf{Cats are very territorial, and they mark their territory by rubbing against objects.}}}\\
&\multicolumn{11}{l}{ They also like to be scratched, which helps them stay healthy. So they rub against your legs}\\
&\multicolumn{11}{l}{to mark you as their territory, and also to get a scratch.}\\
\multirow{2}{*}{\text{CLHA}$_{3}$}&\multicolumn{11}{l}{\faCheck{   }{\textbf{Yes, cats do this to communicate with humans and to mark their territory. They also}}}\\
&\multicolumn{11}{l}{ \textbf{enjoy the sensation of rubbing against something soft and scratchy. }}\\
\multirow{3}{*}{\text{CLHA$_{2}$ w/o $\mathcal{L}_{\text{clha}}$}}&\multicolumn{11}{l}{\faTimes{   }{Cats are very affectionate animals, and they like to be close to their human companions. }}\\
&\multicolumn{11}{l}{  They also like to be scratched, which is why they often rub against your legs.  Cats also like}\\
&\multicolumn{11}{l}{ to be petted, so if you stroke their fur, they’ll be very happy.}\\
\bottomrule
\end{tabular}}
\caption{Case study, cases are sampled from Harmless$_{base}$ and Helpful$_{base}$}
\label{tab:casestudy}
 \end{center}
\end{table*}

In this section, we provide a detailed examination of two cases taken from Harmless$_{base}$ and Helpful$_{base}$ to evaluate the performance of CLHA. Refer to Table~\ref{tab:casestudy} for the original user queries and the respective responses from CLHA$_{2}$ and CLHA$_{3}$. For a more comprehensive analysis, we also include the results after omitting the contrastive learning loss, denoted as CLHA$_{2}$ w/o $\mathcal{L}_{\text{clha}}$.

In the first case, the absence of the contrastive learning loss in the model leads to difficulties in accurately understanding the query. In comparison, CLHA$_{2}$ employs a strategy to avoid harmful queries, aiming to reduce the likelihood of generating unsafe responses. Meanwhile, CLHA$_{3}$ offers a response that directly addresses the unethical intent behind the query ``steal a wallet from a lady'', stressing the legal and ethical implications of such actions. The answer is coherent and ensured the dissemination of accurate information.

For the second case, all three models appears to produce coherent answers. However, a deeper examination reveals differences in their effectiveness. The model without the contrastive learning loss fails to recognize the important concept of ``cats having a sense of territory''. As a result, its response, while coherent, lacks key information. On the other hand, both CLHA models identifies the main point of the query. However, the response from CLHA$_{2}$ is longer than necessary. CLHA$_{3}$, in contrast, provides a concise and accurate response, making it the most informative of the three.


\section{Related Work}

In the realm of artificial intelligence, ensuring that large language models are in sync with human preferences has become a critical area of exploration. Over the past few years, this topic has seen a surge in research efforts and various methodologies have emerged. One particularly notable approach in this direction is the Reinforcement Learning from Human Feedback (RLHF), with pioneering work such as InstructGPT as demonstrated by~\citet{ouyang2022training}. RLHF, although robust in its essence, presents a framework that isn't free from potential hurdles. The method requires massive computational power and intricate setups during the training phase. As pointed out by~\citet{zheng2023secrets}, hyperparameter sensitivity can be a major bottleneck, leading to considerable costs in both monetary and time aspects during the training process.

Recognizing these limitations, the academic community has sought out innovative alternatives that diverge from traditional reinforcement learning (RL) paradigms. One such notable endeavor is by~\citet{rafailov2023direct}, which revisits the optimization objectives of RLHF. Instead of conventional strategies, they suggest direct optimization on preference datasets, a novel approach that simplifies the alignment process. In a similar vein of innovation, some researchers propose harnessing the innate capabilities of models to self-align. A case in point is the CoH (Chain of Hindsight) method posited by~\citet{liu2023chain}. This technique innovatively integrates human feedback with natural language processing. By employing question-answer pairs within a single sentence structure, CoH taps into the model's intrinsic comprehension abilities, facilitating human alignment. This method not only makes the alignment process more manageable but also avoids the pitfalls of high computational demands.
Apart from these, there is still a growing realization in the community about the pivotal role sequence data can play in tasks associated with human alignment. Researches, such as those by~\citet{yuan2023rrhf} and~\citet{song2023preference}, have delved into this domain, introducing intuitive mechanisms like RRHF and PRO. Such explorations are instrumental in broadening our understanding and presenting a more holistic picture of the challenges and potential solutions in the ever-evolving landscape of human-aligned tasks.

\section{Conclusion}

In this paper, we propose a simple yet effective framework CLHA for exploring the alignment of human preferences. First, we have devised a rescoring strategy to eliminate the noise information caused by erroneous samples. This allows us to filter out higher quality positive samples in the dataset, providing a more accurate measure of the degree of human preference between two samples. Second, we have integrated pairwise contrastive loss and adaptive supervised fine-tuning loss to ensure a well-balanced distinction between positive and negative samples, resulting in an enhanced alignment with human preferences. Our experimental results on benchmark datasets demonstrate that CLHA significantly outperforms existing methods.

\section*{Acknowledgements}
This work was supported by National Key Research and Development Program of China (2022YFF0902100), National Natural Science Foundation of China (62376262), Shenzhen Science and Technology Innovation Program (KQTD20190929172835662), Shenzhen Basic Research Foundation (JCYJ20210324115614039). 










\nocite{*}
\section*{Bibliographical References}\label{sec:reference}

\bibliographystyle{lrec-coling2024-natbib}
\bibliography{lrec-coling2024-example}

\label{lr:ref}
\bibliographystylelanguageresource{lrec-coling2024-natbib}
\bibliographylanguageresource{languageresource}

\end{document}